\documentclass[runningheads]{llncs}

\usepackage{array}
\usepackage{booktabs}
\usepackage[T1]{fontenc}
\usepackage{amsmath}
\usepackage{newtxtext}
\usepackage[varvw]{newtxmath}
\usepackage{graphicx}
\usepackage{longtable}
\usepackage{url}

\usepackage{hyperref}

\begin{document}

\title{An Analysis of Role Semantics in Directed Dynamic Graph Modeling for Edge Classification}
\titlerunning{Role Semantics in Dynamic Edge Classification}

\author{Tyler Bonnet \and
Marek Rei}

\authorrunning{T. Bonnet and M. Rei}

\institute{Imperial College London, London, UK\\
\email{\{t.bonnet24, marek.rei\}@imperial.ac.uk}}



\maketitle

\begin{abstract}

Edge classification on directed dynamic graphs is critical for understanding evolving user-item interactions and behavioral patterns in recommender systems and social platforms. Existing dynamic graph architectures typically process these interactions using shared parameters for nodes with semantically distinct roles, including source, destination, query, and neighbor. While abundant supervision in tasks such as link prediction provides sufficient gradient signal for models to implicitly deduce role semantics and overcome this conflation, there are often far fewer labels available for edge classification. We hypothesize that in such regimes, the ability of models to implicitly separate roles is restricted, degrading downstream performance. To empirically investigate this relationship, we introduce an experimental framework for sequence-based models that explicitly disentangles functional roles using targeted embeddings and an asymmetric pretraining objective. By providing architectural priors for these roles, the framework serves as an evaluative vehicle to isolate and measure the resulting performance delta across a range of label budgets. Our empirical evaluation demonstrates that this explicit role disentanglement yields substantial and consistent relative performance gains over baselines. Furthermore, representational analyses reveal that the mechanism by which the framework provides these benefits is a bifurcation of source and destination latent spaces that improves linear separability.

\end{abstract}

\section{Introduction}

Dynamic graphs are widely used to model systems where interactions between entities evolve over time, appearing frequently in domains such as e-commerce and social networks to serve important functions in recommender systems. In practice, predicting the mere occurrence of an interaction is often insufficient. Platforms must also anticipate the specific nature of the engagement, such as whether a consumer will leave a 5-star review or a community member will provide a helpful forum answer. To capture temporal dynamics, research has largely shifted from discrete snapshot-based approaches \cite{sankar2020dysat,pareja2020evolvegcn} to continuous-time dynamic graph networks that process interactions as a continuous chronological sequence \cite{kumar2019jodie,trivedi2019dyrep,rossi2020temporal}. More recently, the field adopted the Transformer \cite{vaswani2017attention} to overcome the memory constraints of recurrent models, using self-attention mechanisms to better encode dependencies in interaction sequences \cite{wang2021tcl,yu2023dygformer}. The field has also developed self-supervised learning methods to improve generalization by pretraining representations via generative and contrastive learning objectives \cite{xu2023cldg,zhu2024idol,gao2025dvgmae,liu2025dygmae}.

Beyond capturing temporal dynamics, modeling interactions in directed dynamic graphs also involves processing distinct functional roles, such as source, destination, query, and neighbor. Existing architectures generally approach this through unified processing, applying identical transformation logic across all roles with shared parameters. This architectural design requires the models to implicitly learn specialized feature mappings for each role. While the abundant structural supervision available in link prediction provides sufficient gradient signal to deduce these role semantics and overcome conflation, edge classification often relies on a limited number of annotated labels. We hypothesize that in such regimes, the ability of the model to implicitly separate roles is restricted, degrading downstream performance.

To investigate the effects of this role-semantic conflation, we propose Dynamic Graph Node-Role-Oriented Latent Encoding (DyGnROLE)\footnote{Code will be made available upon publication.}, an experimental framework for sequence-based models designed to explicitly disentangle functional roles for dynamic edge classification. We encode this role awareness through three architectural priors: (1) role-semantic positional encodings that distinguish query and neighbor roles across both source and destination contexts; (2) within- and cross-sequence frequency embeddings to capture role-specific recurring interactions and local connectivity patterns; and (3) dual-CLS representation pooling to extract separate global representations for the source and destination nodes. To organize these separated representation spaces, we use directional role alignment (DRA), an asymmetric contrastive pretraining objective. By treating source representations as queries to retrieve destination targets, DRA provides relational cues that structure the latent space around the directional interaction process itself. The objective also uses a historical positive masking strategy to prevent established relationships from being penalized as false negatives during in-batch comparisons. DyGnROLE essentially serves as an experimental control, allowing us to analyze the effect of relieving the network of implicit role modeling on downstream edge classification performance across a range of label budgets.

For our analysis, we integrate the Transformer-based DyGFormer \cite{yu2023dygformer} architecture into the DyGnROLE framework to evaluate its performance on five datasets from the Dynamic Text-Attributed Graph Benchmark (DTGB) \cite{zhang2024dtgb} with and without explicit role modeling. We also compare these results against several recent supervised and self-supervised dynamic graph baselines. Our empirical evaluation demonstrates that explicit role disentanglement yields substantial and consistent performance improvements. Representational analyses further reveal that the framework achieves these gains by bifurcating source and destination latent spaces to prevent parameter conflation and improve linear separability. Finally, an ablation study isolates the impact of each architectural component and the pretraining strategy, confirming exactly how relieving the network of implicit role modeling affects downstream edge classification performance.

\section{Related Work}
\label{sec:related_work}

We review existing approaches to dynamic graph representation learning, focusing on prevailing architectural design choices and self-supervised pretraining strategies.

\subsection{Role-Agnostic Processing in Dynamic Graph Models}

To address broad expressivity challenges, existing dynamic graph architectures generally exchange structural role-awareness for parameter efficiency. Unified encoding schemes are standard across memory-based models \cite{trivedi2019dyrep}, attention-based architectures \cite{xu2020tgat}, MLP-based networks \cite{cong2023graphmixer}, and walk-based approaches \cite{wang2021cawn}. This role-agnostic processing extends to state-of-the-art Transformer-based models, including DyGFormer \cite{yu2023dygformer} and TCL \cite{wang2021tcl}. Because these Transformer architectures are permutation-invariant and omit role-semantic positional encodings, their encoding streams apply the same parameterized transformations to both interacting nodes. Even though TCL uses a two-stream encoder to partition source and destination subgraphs, it applies identical co-attention weights to both streams. DyGFormer encodes neighbor frequency information using a shared co-occurrence MLP for both source and destination histories. In these symmetric designs, directionality is typically imposed via a fixed concatenation order at the final classification layer (DyGFormer), or managed through a mathematically commutative scoring function (TCL).

While shared parameterization has been effective for link prediction, where abundant supervision allows networks to implicitly deduce distinct role behaviors, it remains an open question whether this implicit inference is sufficient for edge classification where there is often far less supervision. We investigate this question with the DyGnROLE framework, which allows us to evaluate the impact of explicit role modeling with dedicated learnable embeddings and the DRA objective during a pretraining phase.

\subsection{Self-Supervised Learning for Dynamic Graphs}

To address generalization challenges, research has turned to self-supervised learning to extract training signals from unlabeled data. Contrastive methods such as CLDG \cite{xu2023cldg} and IDOL \cite{zhu2024idol} construct multiple views based on chronological evolution. CLDG samples distinct timespan windows to enforce temporal translation invariance across the history of a node, while IDOL uses topology-monitorable sampling to ensure representation coherence between current and historical embeddings. Generative methods such as DyGMAE \cite{liu2025dygmae} and DVGMAE \cite{gao2025dvgmae} use masked autoencoding to capture the temporal evolution of graphs. These models mask structural and temporal information within historical interactions and learn representations by reconstructing the missing data, with DVGMAE incorporating variational inference to model interaction uncertainty.

While optimizing for invariance or reconstruction allows these models to achieve identity stability and structural recovery, they remain inherently symmetric and lack explicit role priors. It remains unexplored how self-supervised learning can be designed around the directional asymmetry of real-world interactions. We address this with the DRA objective, which formulates pretraining as a source-to-destination retrieval task with historical positive masking.

\section{Method}
\label{sec:method}

\begin{figure*}[ht]
    \centering
    \includegraphics[width=\textwidth]{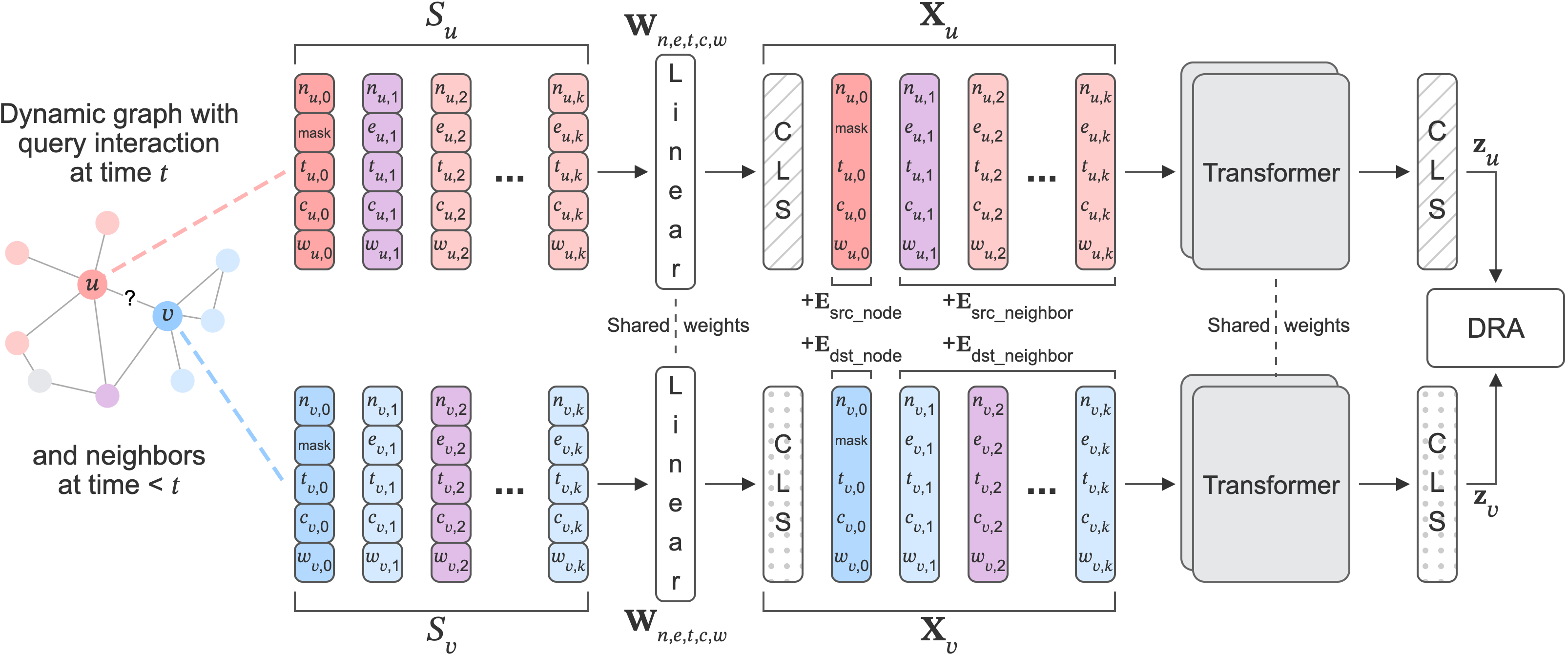}
    \caption{\textbf{Overview of the DyGnROLE Framework.} \textbf{(Left)} For a query interaction between nodes $u$ and $v$ at time $t$, historical one-hop neighbors are retrieved. $u$ and $v$ combined with their respective neighbors form node sequences $S_u$ and $S_v$. For each event in these sequences, its node feature $n$, edge feature $e$, interaction time delta $t$, cross-sequence neighbor frequency $c$, and within-sequence neighbor frequency $w$ are extracted, and the edge feature of the query nodes is masked. \textbf{(Center)} All five extracted features are first independently projected via separate linear layers $\mathbf{W}_{n,e,t,c,w}$ and concatenated. role-semantic positional encodings (RSPE) are added to the projected features to distinguish source query, destination query, source neighbors, and destination neighbors. Distinct CLS tokens are then prepended to form the final augmented input matrices $\mathbf{X}_u$ and $\mathbf{X}_v$. The framework accommodates sequence models. In our study, we integrate DyGFormer with its Transformer (Pre-LN, GELU) architecture as the backbone encoder. \textbf{(Right)} The sequences $\mathbf{X}_u$ and $\mathbf{X}_v$ are processed independently by the Transformer encoder, using shared weights. The CLS tokens are extracted to serve as the final source and destination representations $\mathbf{z}_u$ and $\mathbf{z}_v$, which are optimized via the directional role alignment (DRA) loss.}
    \label{fig:dygnrole_architecture}
\end{figure*}

\subsection{Problem Formulation}

A dynamic graph is defined as $\mathcal{G} = (\mathcal{V}, \mathcal{E}, \mathcal{T}, \mathcal{D}, \mathcal{R})$ where $\mathcal{V}$ is the node set, $\mathcal{E} \subseteq \mathcal{V} \times \mathcal{V} \times \mathcal{T}$ contains timestamped edges, $\mathcal{T}$ is the set of timestamps at which edges occur, $\mathcal{D}$ is the set of node attributes, and $\mathcal{R}$ is the set of edge attributes. An edge $(u, v, t) \in \mathcal{E}$, with its associated node and edge attributes, connects a source node $u$ to a destination node $v$ at time $t$, and has a label $y_{uv}^t$ representing the edge class. Given a set of historical edges up to $t$, defined as $\mathcal{E}_{\leq t} = \{(u, v, t', y_{uv}^{t'}) \in \mathcal{E} \mid t' \leq t\}$, the objective is to learn a function $f: (\mathcal{V}, \mathcal{E}_{\leq t}, \mathcal{D}, \mathcal{R}) \rightarrow y_{uv}^{t+1}$ that predicts the class of an interaction at $t+1$.

\subsection{Sequence Construction and Base Features}
\label{sec:sequence_construction}

To construct the inputs for a target source node $u$ and destination node $v$ at time $t$, the model samples their respective one-hop temporal neighbors. These historical sequences are truncated or padded to a fixed length $k$, and the query nodes $u$ and $v$ are prepended to form the final interaction sequences $S_u$ and $S_v$. For any given event $i$ in these sequences (e.g., $i \in S_u$), the framework extracts a discrete set of features: the node text embedding $n_{u,i}$, edge text embedding $e_{u,i}$, temporal delta $t_{u,i}$, cross-sequence frequency $c_{u,i}$, and within-sequence frequency $w_{u,i}$. To prevent target leakage, the edge text embedding for the query nodes themselves is masked. Finally, these five extracted features are projected through independent linear layers $\mathbf{W}_{n,e,t,c,w}$ and concatenated, establishing the base continuous representations for the events in $S_u$ and $S_v$.

\subsection{Architectural Priors for Explicit Role Modeling}
\label{sec:model_architecture}

Building upon these base event representations, the framework introduces three architectural components designed to explicitly disentangle role semantics.

\subsubsection{Role-Semantic Positional Encodings (RSPE)}

Because Transformers are inherently permutation-invariant, passing raw concatenated features through the encoder forces the model to treat source and destination sequences identically. To distinguish source from destination and query from neighbor, we inject learnable role identifiers directly into the latent space. We define four distinct learnable embeddings with dimension $D = 250$: $\mathbf{E}_{\text{src\_node}}$, $\mathbf{E}_{\text{src\_neighbor}}$, $\mathbf{E}_{\text{dst\_node}}$, and $\mathbf{E}_{\text{dst\_neighbor}}$. These embeddings are added to the projected base features to encode the role of the node in a given interaction sequence.

\subsubsection{Neighbor Frequency Embeddings}

To capture local connectivity patterns, we integrate a module that maps embeddings to node frequencies within and across sequences. For event $i$ in source sequence $S_u$, it calculates the within-sequence frequency $w_{u,i}$ (appearances in $S_u$) and cross-sequence frequency $c_{u,i}$ (appearances in $S_v$) of the participating node, applied symmetrically for $S_v$. The embedding tables the counts are mapped to are separate for source and destination. They replace the role-agnostic co-occurrence encoder in DyGFormer \cite{yu2023dygformer} (further integration details provided in Section \ref{sec:backbones}).

To prevent overfitting to rare values, only counts occurring at least $N_{\min}=10{,}000$ times during pretraining receive distinct indices, while rarer counts map to an UNK token. Formally, the index set is defined as $\mathcal{V}_{\text{embed}} = \{ c \mid N_c \geq N_{\min} \}$, where $N_c$ denotes the occurrence of count $c$. The embedding for a count $c$ is defined as $\mathbf{e}_c = \text{Embedding}(c)$ if $c \in \mathcal{V}_{\text{embed}}$, and $\text{Embedding}(\text{UNK})$ otherwise, yielding $\mathbf{e}_c \in \mathbb{R}^{d_c}$.

\subsubsection{Dual-CLS Representation Pooling}

The framework prepends distinct, learnable CLS tokens to the source and destination sequences prior to encoding. While applying mean-pooling independently to the source and destination branches yields role-specific aggregations, dedicated CLS tokens act as parameterized attention sinks that can selectively gather relevant context without the algebraic restrictions of a uniform average. These tokens are designed to aggregate the entire interaction context and extract separate global representations for the source and destination nodes. After processing through the backbone Transformer, these distinct tokens are extracted to serve as the final representations $\mathbf{z}_u$ and $\mathbf{z}_v$ for the DRA pretraining objective.

\subsection{Directional Role Alignment (DRA)}
\label{sec:dra}

A key challenge in applying noise-contrastive estimation \cite{oord2018representation} to dynamic graphs is negative sampling. Unlike prior works that penalize historical positives \cite{rossi2020temporal,poursafaei2022edgebank}, we use in-batch negative sampling paired with historical positive masking. We query the neighbor sampler up to the batch minimum timestamp $t_{min}$ to construct a binary mask $M \in \{0,1\}^{N \times N}$. If destination $j$ is a historical neighbor of source $i$, $M_{ij} = 0$, and their similarity score is set to $-\infty$ before softmax. This preserves established structural relationships while allowing recurring interactions to provide positive signals. The loss is formalized as:
$$
\mathcal{L}_{\mathrm{DRA}}
= -\frac{1}{N} \sum_{i=1}^{N} \log \frac{\exp\!\left(\frac{\mathbf{z}_u^{(i)} \cdot \mathbf{z}_v^{(i)}}{\tau}\right)}{\sum_{j=1}^{N} M_{ij} \exp\!\left(\frac{\mathbf{z}_u^{(i)} \cdot \mathbf{z}_v^{(j)}}{\tau}\right)},
$$
where $N$ is the batch size, $\mathbf{z}_u$ and $\mathbf{z}_v$ are the $\ell_2$-normalized representations, and $\tau = 0.07$. The mask $M_{ij} = 1$ if $i = j$ or if $j$ has never interacted with $i$, otherwise $M_{ij} = 0$.

While the architectural priors provide the capacity for asymmetric role modeling, DRA supplies the organizing training signal. Source and destination histories are processed independently into distinct latent spaces. The loss is computed row-wise over a source-to-destination similarity matrix, treating source embeddings as queries to retrieve destination targets. By restricting supervision to this direction, the model learns a role-dependent compatibility function rather than a symmetric node similarity, which is critical for directed networks where roles exhibit distinct interaction patterns.

Finally, historical positive masking introduces temporal asymmetry. Nodes initially serve as negative candidate pairs, but following their first interaction, they are permanently removed from future negative comparisons and contribute only as positive pairs during subsequent interactions. This monotonic progression from unseen to observed ensures that established history is not corrupted by arbitrary negative sampling.

\subsection{Sequence Transformer Integration}
\label{sec:backbones}

We integrate DyGFormer \cite{yu2023dygformer} into the DyGnROLE framework with adaptations that focus on substituting baseline components that algebraically conflate role signals, while preserving the core Transformer encoder. To align this architecture with the framework, the input space is expanded from 4 to 5 modular channels. Independent processing of the source and destination branches is enforced during pretraining to prevent contrastive leakage, with joint attention restored for finetuning.

\subsection{Pretraining and Finetuning Protocol}
\label{sec:training_procedure}

During pretraining, models integrated into the DyGnROLE framework optimize the DRA objective, while self-supervised baseline models optimize their native objectives using the most recent 500,000 edges of the training split. Validation is performed on the most recent 30,000 edges of the validation split. Pretraining proceeds until convergence, governed by early stopping based on validation performance with a patience of 10 epochs.

For the downstream edge classification task, both the pretrained backbone and a randomly initialized linear classification layer are finetuned end-to-end to optimize a standard cross-entropy loss between the predicted logits and the ground-truth edge labels. Finetuning is evaluated across training label budgets ranging from 1,000 to 20,000, selecting the most recent edges from the training split for each budget. A corresponding validation set is constructed using the most recent edges from the validation split, sized at 15\% of the active training budget. Similar to pretraining, finetuning runs until convergence, relying on early stopping with a patience of 20 epochs based on validation performance.

\section{Experiments}
\label{sec:experiments}

To empirically evaluate the impact of explicitly disentangling functional roles, we conduct a series of experiments comparing the DyGnROLE framework to existing supervised and self-supervised dynamic graph models. The evaluation has four core objectives: assessing overall edge classification performance, analyzing model robustness under varying degrees of label availability, isolating the contributions of individual architectural components and the DRA pretraining objective through an ablation study, and examining the geometry of the learned latent spaces through representational analyses.

\subsection{Datasets and Tasks}
\label{sec:datasets}

We evaluate models on five datasets from the Dynamic Text-Attributed Graph Benchmark (DTGB) \cite{zhang2024dtgb}. DTGB is selected because its datasets provide explicit edge labels necessary for classification, whereas many standard dynamic graph datasets (e.g., MOOC, LastFM, or Wikipedia) are restricted to link prediction. The datasets are available through the official DTGB repository\footnote{\url{https://github.com/zjs123/DTGB}}.

\begin{table*}[t]
\centering
\small
\setlength{\tabcolsep}{3.5pt}
\caption{Edge classification results with 10k training labels showing average Macro F1 score percent $\pm$ standard deviation. \textbf{Bold} indicates the best result on each dataset. Relative performance gains from DyGnROLE over the original DyGFormer architecture are shown in green superscript. Statistical significance of DyGnROLE over the strongest baseline is indicated by $^{\ast} p < 0.10$, $^{\dagger} p < 0.05$, and $^{\ddagger} p < 0.01$ under a two-tailed paired $t$-test for individual datasets and Wilcoxon signed-rank test for the average.}
\label{tab:results_table_sig_10k}
\resizebox{0.9\textwidth}{!}{%
\begin{tabular}{lcccccc}
\toprule
\textbf{Method} & \textbf{Stack E} & \textbf{Stack U} & \textbf{Amazon} & \textbf{Google} & \textbf{Yelp} & \textbf{Avg} \\
\midrule
Random & $46.74_{\pm 0.14}$ & $44.74_{\pm 0.05}$ & $15.22_{\pm 0.03}$ & $15.02_{\pm 0.07}$ & $16.93_{\pm 0.02}$ & $27.73_{\pm 0.06}$ \\
Majority & $42.82_{\pm 0.00}$ & $44.65_{\pm 0.00}$ & $15.76_{\pm 0.00}$ & $15.93_{\pm 0.00}$ & $13.99_{\pm 0.00}$ & $26.63_{\pm 0.00}$ \\
\midrule
DyRep & $43.52_{\pm 1.29}$ & $45.08_{\pm 0.76}$ & $15.00_{\pm 2.35}$ & $14.54_{\pm 2.35}$ & $14.80_{\pm 1.94}$ & $26.59_{\pm 1.74}$ \\
TCL & $42.82_{\pm 0.00}$ & $44.65_{\pm 0.00}$ & $15.76_{\pm 0.00}$ & $15.93_{\pm 0.00}$ & $13.99_{\pm 0.00}$ & $26.63_{\pm 0.00}$ \\
IDOL & $42.82_{\pm 0.00}$ & $44.65_{\pm 0.00}$ & $15.76_{\pm 0.00}$ & $15.93_{\pm 0.00}$ & $14.13_{\pm 0.28}$ & $26.66_{\pm 0.06}$ \\
DyGMAE & $42.82_{\pm 0.00}$ & $44.65_{\pm 0.00}$ & $15.76_{\pm 0.00}$ & $15.93_{\pm 0.00}$ & $14.20_{\pm 0.42}$ & $26.67_{\pm 0.08}$ \\
DVGMAE & $42.82_{\pm 0.00}$ & $45.76_{\pm 2.11}$ & $16.61_{\pm 0.74}$ & $15.93_{\pm 0.00}$ & $14.04_{\pm 0.04}$ & $27.03_{\pm 0.58}$ \\
DyGFormer & $42.82_{\pm 0.00}$ & $44.65_{\pm 0.00}$ & $15.76_{\pm 0.00}$ & $15.93_{\pm 0.00}$ & $20.45_{\pm 5.59}$ & $27.92_{\pm 1.12}$ \\
JODIE & $44.21_{\pm 2.43}$ & $45.90_{\pm 1.99}$ & $17.69_{\pm 1.18}$ & $17.73_{\pm 0.96}$ & $17.42_{\pm 1.72}$ & $28.59_{\pm 1.65}$ \\
CLDG & $42.82_{\pm 0.00}$ & $44.65_{\pm 0.00}$ & $15.76_{\pm 0.00}$ & $15.93_{\pm 0.00}$ & $24.08_{\pm 1.04}$ & $28.65_{\pm 0.21}$ \\
GraphMixer & $52.45_{\pm 7.88}$ & $44.65_{\pm 0.00}$ & $17.09_{\pm 2.67}$ & $15.93_{\pm 0.00}$ & $16.06_{\pm 1.49}$ & $29.24_{\pm 2.41}$ \\
CAWN & $42.81_{\pm 0.00}$ & $44.65_{\pm 0.00}$ & $18.01_{\pm 2.39}$ & $17.79_{\pm 1.83}$ & $25.72_{\pm 2.05}$ & $29.80_{\pm 1.25}$ \\
TGAT & $42.81_{\pm 0.00}$ & $44.65_{\pm 0.00}$ & $21.50_{\pm 1.46}$ & $17.42_{\pm 2.96}$ & $23.87_{\pm 8.08}$ & $30.05_{\pm 2.50}$ \\
\midrule
DyGnROLE & {\boldmath $55.56^{{\color{green!60!black}+29.8\%}}_{\pm 0.38}$} & {\boldmath $53.94^{{\color{green!60!black}+20.8\%}^{\ddagger}}_{\pm 0.70}$} & {\boldmath $22.97^{{\color{green!60!black}+45.8\%}}_{\pm 1.95}$} & {\boldmath $22.40^{{\color{green!60!black}+40.6\%}^{\ddagger}}_{\pm 1.35}$} & {\boldmath $32.95^{{\color{green!60!black}+61.2\%}^{\ddagger}}_{\pm 0.90}$} & {\boldmath $37.56^{{\color{green!60!black}+34.5\%}^{\ddagger}}_{\pm 1.05}$} \\
\bottomrule
\end{tabular}%
}
\end{table*}

\textbf{Amazon, Google, and Yelp:} These e-commerce datasets model consumer behavior and user satisfaction. The Amazon dataset is based on Amazon review data for Movies and TV products\footnote{\url{https://cseweb.ucsd.edu/~jmcauley/datasets/amazon_v2/}}. The Google dataset uses Google Local review data covering Connecticut businesses\footnote{\url{https://mcauleylab.ucsd.edu/public_datasets/gdrive/googlelocal/}}. The Yelp dataset uses the Yelp Open Dataset to model reviews of businesses such as restaurants and hotels\footnote{\url{https://www.yelp.com/dataset}}. Nodes represent users and businesses or products. Edges represent a user reviewing an item. The node attributes include user profiles, product names, categories, and descriptions. The edge texts capture the raw written user reviews. On these datasets, the task of \textbf{rating prediction} is performed, where the model predicts a future rating of 1 to 5 stars assigned by the user for a product or service.

\textbf{Stack E and Stack U:} Sourced from Stack Exchange\footnote{\url{https://archive.org/details/stackexchange}}, these forums focus on electrical engineering and the Ubuntu operating system, respectively. Nodes represent users or questions. Edges indicate the submission of an answer to a question or a comment. Node texts include user profiles (name, location, introduction) or question titles and bodies. Edge texts contain the raw written answer or comment. The task performed on these datasets is \textbf{post usefulness prediction}, where the model predicts whether a future reply of a user will be ``useful'' or ``useless'' determined by the community score (upvotes minus downvotes, where a score greater than 1 is useful).

\begin{figure}[t]
    \centering
    \includegraphics[width=0.72\columnwidth]{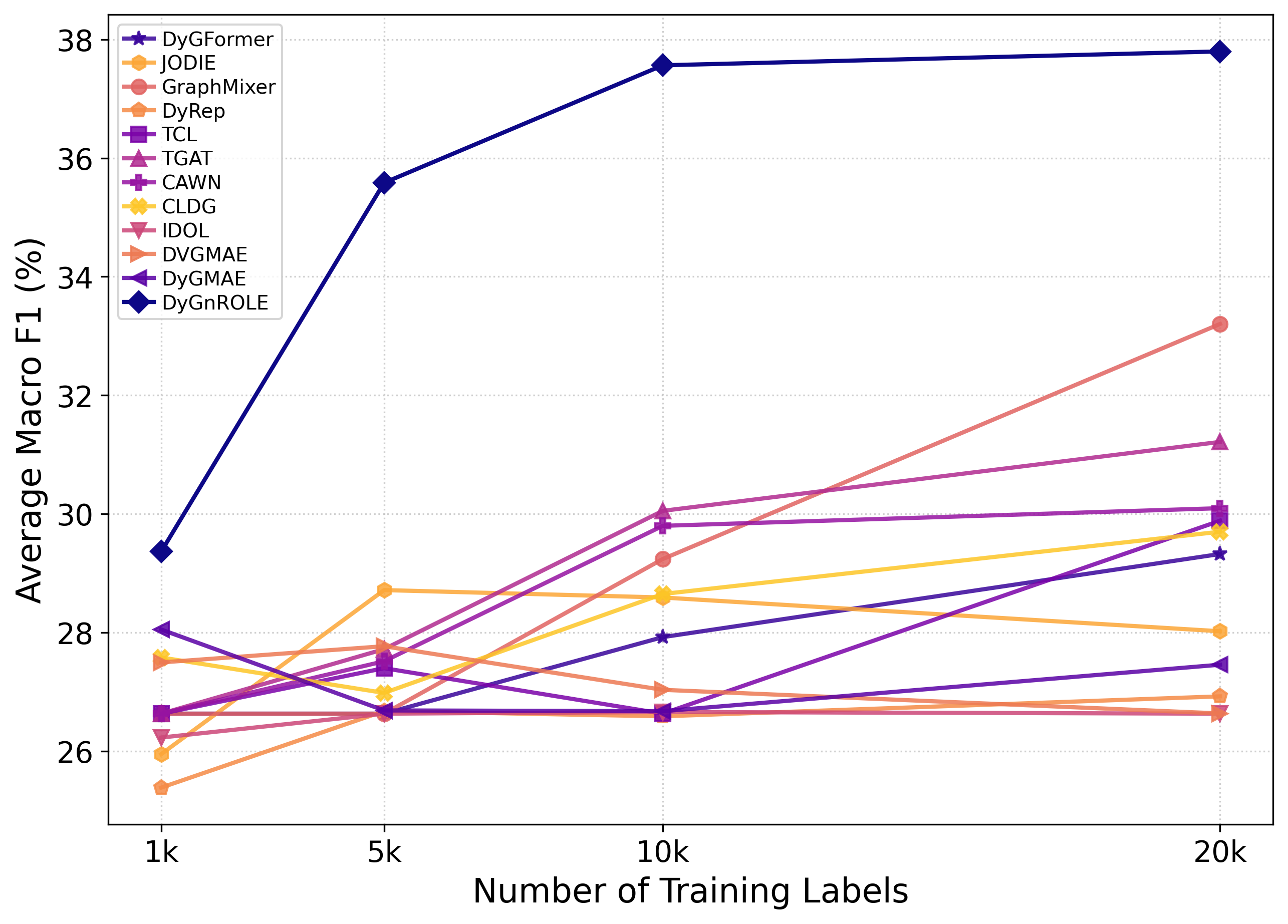}
    \caption{Average Macro F1 scores across a range of training label budgets from 1k to 20k.}
\label{fig:performance_vs_scale}
\end{figure}

\subsection{Baselines}
\label{sec:baselines}

We evaluate a range of supervised and self-supervised models. The supervised class includes memory-based models \cite{kumar2019jodie,trivedi2019dyrep}, a walk-based model \cite{wang2021cawn}, and attention-based models \cite{xu2020tgat,wang2021tcl,cong2023graphmixer,yu2023dygformer}. The self-supervised class includes contrastive models \cite{xu2023cldg,zhu2024idol} and generative masked autoencoders \cite{liu2025dygmae,gao2025dvgmae}.

\subsection{Main Results}
\label{sec:results}

The results in Table \ref{tab:results_table_sig_10k} demonstrate that models relying on shared parameters exhibit constrained performance without abundant structural supervision. Several baselines collapse to majority-class prediction across multiple datasets, falling back on the highly skewed class distributions. Integrating the base DyGFormer architecture into the DyGnROLE framework yields an average Macro F1 score of 37.56\%, a statistically significant 34.5\% relative improvement. This performance difference is particularly pronounced within the e-commerce domain, where integration provides relative gains ranging from 40.6\% to 61.2\%.

Evaluating performance across a range of label budgets (Figure \ref{fig:performance_vs_scale}) reveals divergent trajectories. Most baseline architectures show limited capacity at lower budgets, demonstrating improvements as they approach 20,000 labels. DyGnROLE exhibits a different trajectory, capturing the training signal earlier in the scaling process. However, at the 1,000 label budget, the margin between the baselines and the framework narrows. This reduced relative gain warrants a closer examination of how individual architectural components behave under the lowest label budgets, which we explore further in Section \ref{sec:ablation}. Overall, these results support the hypothesis that relieving a network of implicit role modeling alters its data requirements for dynamic edge classification.

\subsection{Ablation Study}
\label{sec:ablation}

\begin{table*}[t]
\centering
\small
\setlength{\tabcolsep}{3.5pt}
\caption{Ablation results showing average Macro F1 score percent $\pm$ standard deviation. \textbf{Bold} indicates the best result on each dataset. Baseline w/o DRA denotes original DyGFormer and w/ DRA denotes DyGFormer adapted for DRA pretraining by encoding source and destination node sequences separately during pretraining and jointly during finetuning following the DyGnROLE protocol. The remaining variants in the DyGnROLE section are remove-one ablations, removing one architectural component at a time from the DyGnROLE framework.}
\label{tab:ablation}
\resizebox{0.75\textwidth}{!}{%
\begin{tabular}{lcccccc}
\toprule
\textbf{Variants} & \textbf{Stack E} & \textbf{Stack U} & \textbf{Amazon} & \textbf{Google} & \textbf{Yelp} & \textbf{Avg} \\
\midrule
\multicolumn{7}{l}{\textit{Baseline}} \\
\midrule
w/o DRA & $42.82_{\pm 0.00}$ & $44.65_{\pm 0.00}$ & $15.76_{\pm 0.00}$ & $15.93_{\pm 0.00}$ & $20.45_{\pm 5.59}$ & $27.92_{\pm 1.12}$ \\
w/ DRA & {\boldmath $56.03_{\pm 1.40}$} & {\boldmath $55.03_{\pm 0.86}$} & $20.73_{\pm 1.71}$ & $16.89_{\pm 1.90}$ & $28.61_{\pm 3.48}$ & $35.46_{\pm 1.87}$ \\
\midrule
\multicolumn{7}{l}{\textit{DyGnROLE}} \\
\midrule
w/o RSPE & $54.32_{\pm 1.77}$ & $51.89_{\pm 3.51}$ & $22.31_{\pm 0.61}$ & $21.52_{\pm 2.90}$ & $27.88_{\pm 6.99}$ & $35.58_{\pm 3.16}$ \\
w/o CLS & $55.78_{\pm 1.14}$ & $53.70_{\pm 0.99}$ & $21.15_{\pm 2.78}$ & $22.05_{\pm 1.03}$ & $27.82_{\pm 7.22}$ & $36.10_{\pm 2.63}$ \\
w/o NFE & $55.94_{\pm 2.16}$ & $53.31_{\pm 3.54}$ & $22.68_{\pm 1.22}$ & $17.32_{\pm 1.66}$ & $32.78_{\pm 1.56}$ & $36.41_{\pm 2.03}$ \\
Complete & $55.56_{\pm 0.38}$ & $53.94_{\pm 0.70}$ & {\boldmath $22.97_{\pm 1.95}$} & {\boldmath $22.40_{\pm 1.35}$} & {\boldmath $32.95_{\pm 0.90}$} & {\boldmath $37.56_{\pm 1.05}$} \\
\bottomrule
\end{tabular}%
}
\end{table*}

The ablation analysis presented in Table \ref{tab:ablation} isolates the contribution of the DRA objective and the architectural priors. Training the baseline architecture, DyGFormer, with DRA provides the most substantial performance increase, elevating the average Macro F1 from 27.92\% to 35.46\%. This confirms that structuring the latent space around the directional interaction process supplies a critical alignment signal that unified processing lacks. Removing individual architectural components from the complete DyGnROLE framework results in consistent performance degradation, particularly on the e-commerce datasets. Removing RSPE drops average performance to 35.58\%, highlighting the necessity of explicitly distinguishing source query, destination query, source neighbor, and destination neighbor in the latent space. Removing the dual CLS tokens reduces the average score to 36.10\%, causing a sharp decline on the Yelp dataset from 32.95\% to 27.82\%, which indicates that extracting global representations via dedicated tokens provides a cleaner contrastive alignment signal than mean pooling. Removing NFE causes severe degradation on the Google dataset, reducing the score from 22.40\% to 17.32\%. 

While the complete framework achieves the highest overall average, the DRA-augmented baseline slightly outperforms the complete model on the Stack Exchange datasets. To investigate this behavior, we compared the performance of both variants across a range of label budgets, segregated by domain. At the most constrained budget of 1,000 labels, the complete DyGnROLE framework maintains a performance advantage on the Stack Exchange datasets (48.4\% vs 45.1\%). However, as the label budget scales to 5,000 and beyond, the DRA-only variant overtakes the complete framework on these forums. In contrast, on the e-commerce networks, the complete DyGnROLE framework slightly outperforms the DRA-only variant at the 1,000 label budget and provides substantial performance benefits across all subsequent data regimes. This suggests that the benefits of the architectural priors for explicit role modeling vary across network topologies and can lead to overfitting once sufficient label signal becomes available.

\subsection{Pairwise Representational Similarity Analysis}
\label{sec:similarity_analysis}

To verify whether the DRA pretraining objective successfully structures the latent space around the interaction process, we evaluate the pairwise cosine similarity of the pretrained embeddings. Specifically, we measure the alignment margin ($\Delta$) between node pairs that have interacted ($\ge 1$ times) versus pairs that have never interacted ($0$ times). A pretraining objective that maps interacting source and destination nodes closer together in the latent space than non-interacting pairs will yield a positive $\Delta$. To isolate the impact of the pretraining objective of DyGnROLE, we evaluate the w/ DRA ablation variant (baseline architecture with no RSPE, NFE, or CLS), against four existing self-supervised dynamic graph models (CLDG, IDOL, DVGMAE, and DyGMAE) across the first 10,000 edges of the test split for each dataset.

\begin{figure}[t]
    \centering
    \includegraphics[width=\textwidth]{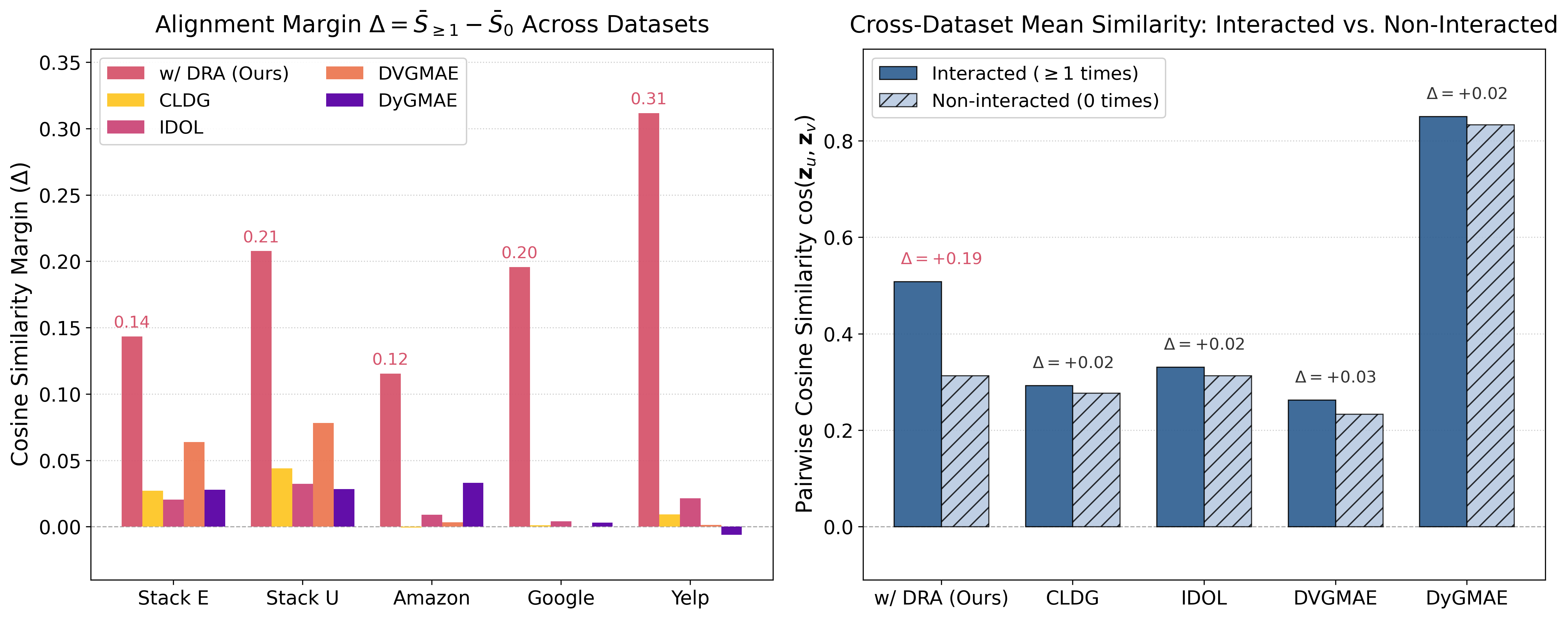}
    \caption{Pairwise representational alignment margins ($\Delta$) across datasets (left) and the cross-dataset mean similarity decomposed by interaction status (right).}
\label{fig:pairwise_cosine_delta}
\end{figure}

The empirical results illustrated in Figure \ref{fig:pairwise_cosine_delta} demonstrate that existing self-supervised methods struggle to differentiate true interacting pairs from non-interacting pairs. Across all five datasets, the baseline models exhibit near-zero alignment margins. For example, generative methods such as DyGMAE produce highly similar embeddings regardless of interaction history ($\Delta = -0.006$ on Yelp and $+0.003$ on Google), while contrastive methods such as CLDG also struggle to establish a discriminative gap ($\Delta = -0.001$ on Amazon). This margin collapse indicates that without an asymmetric objective to guide the representational geometry, symmetric pretraining strategies map interacting and non-interacting nodes into largely overlapping spaces.

In contrast, the w/ DRA model consistently establishes a pronounced alignment margin across all domains. The directional retrieval task, combined with historical positive masking, successfully pulls the latent representations of interacting pairs closer together than those of non-interacting pairs. These geometric properties confirm that the DRA objective supplies the critical gradient signal required to structurally align the latent space around the directed interaction process.

\subsection{Latent Space Visualization and Role Disentanglement}
\label{sec:latent_visualization}

\begin{figure}[h!]
    \centering
    \includegraphics[width=\textwidth]{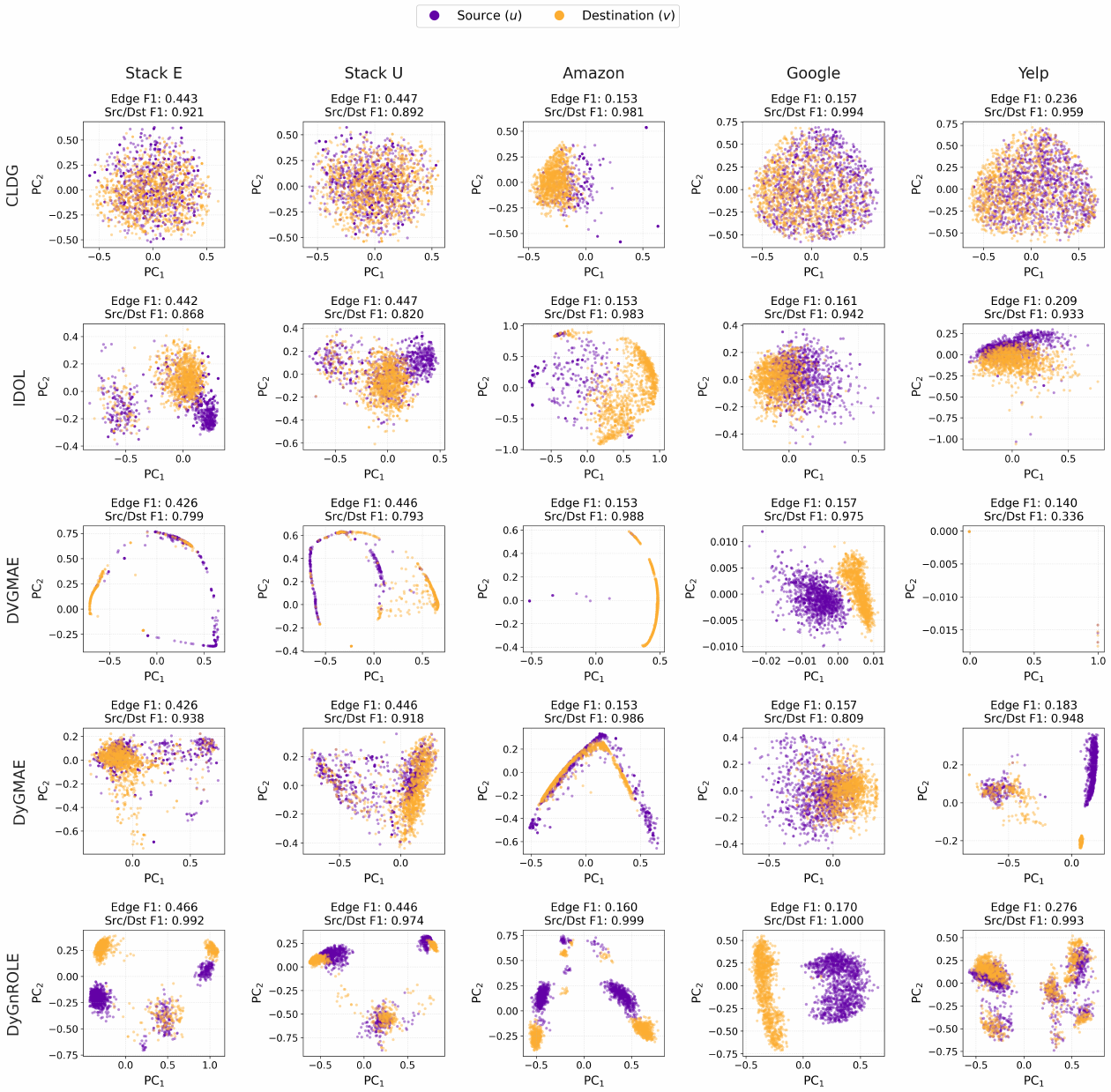}
    \caption{PCA projections of learned latent spaces across five DTGB datasets, colored by node role (source $u$ vs. destination $v$). Linear probing scores evaluate the linear separability of the edge classes (Edge F1) and the source and destination roles (Src/Dst F1).}
\label{fig:pca_role}
\end{figure}

To observe how explicit architectural priors affect the geometric distribution of node representations, we project the learned high-dimensional embeddings into a 2D space using Principal Component Analysis (PCA). Figure \ref{fig:pca_role} illustrates these projections across five DTGB datasets, colored by role (source versus destination).

The visualizations reveal that symmetric contrastive baselines, such as CLDG, project source and destination histories into an overlapping, undifferentiated manifold. Generative approaches such as DVGMAE and DyGMAE often exhibit severe dimensional collapse or form tight geometries that do not reliably separate based on the role of the node. In contrast, the DyGnROLE framework enforces a geometric bifurcation, driving the source and destination representations into distinct submanifolds. This visual disentanglement is supported quantitatively by the linear probing scores (Src/Dst F1) embedded in the figure, where DyGnROLE achieves the highest level of linear separability ($\ge 0.97$) across all evaluated datasets. By separating the role spaces, the framework prevents the parameter conflation that constrains symmetric models.

\section{Conclusion}
\label{sec:conclusion}

In this work, we investigated the impact of shared parameterization for distinct functional roles in directed dynamic graphs by introducing DyGnROLE, an experimental framework designed to explicitly disentangle such roles. By isolating role semantics through targeted positional encodings and frequency embeddings, and organizing the latent space via the asymmetric DRA pretraining objective, the framework served as a structural control to investigate the behavior of dynamic graph sequence models performing the task of edge classification under varying label budgets. Our empirical evaluation demonstrates that this explicit role disentanglement yields substantial and consistent performance improvements over standard architectures. Furthermore, representational analyses reveal that the framework achieves these gains by bifurcating source and destination latent spaces, preventing the parameter conflation that restricts symmetric models.
    
\begin{credits}

\subsubsection{\discintname}
The authors have no competing interests to declare that are relevant to the content of this article.
\end{credits}

\bibliographystyle{splncs04}
\bibliography{references}

\begin{thebibliography}{10}
\providecommand{\url}[1]{\texttt{#1}}
\providecommand{\urlprefix}{URL }
\providecommand{\doi}[1]{https://doi.org/#1}

\bibitem{cong2023graphmixer}
Cong, W., Zhang, S., Kang, J., Yuan, B., Wu, H., Zhou, X., Tong, H., Mahdavi,
  M.: Do we really need complicated model architectures for temporal networks?
  In: International Conference on Learning Representations (ICLR) (2023)

\bibitem{gao2025dvgmae}
Gao, M., Zhang, X., Jiao, P., Li, T., Zhao, Z.: Dvgmae: Self-supervised dynamic
  variational graph masked autoencoder. IEEE Transactions on Neural Networks
  and Learning Systems  \textbf{36}(10),  17901--17913 (2025)

\bibitem{kumar2019jodie}
Kumar, S., Zhang, X., Leskovec, J.: Predicting dynamic embedding trajectory in
  temporal interaction networks. In: Proceedings of the 25th ACM SIGKDD
  International Conference on Knowledge Discovery \& Data Mining. pp.
  1269--1278. ACM, Anchorage, AK, USA (Jul 2019)

\bibitem{liu2025dygmae}
Liu, W., Cheng, J., Pan, Z., He, C., Guan, Q.: Dygmae: A novel dynamic graph
  masked autoencoder for link prediction. In: Chiappa, S., Magliacane, S.
  (eds.) Proceedings of the Forty-first Conference on Uncertainty in Artificial
  Intelligence. Proceedings of Machine Learning Research, vol.~286, pp.
  2685--2700. PMLR (21--25 Jul 2025)

\bibitem{oord2018representation}
Oord, A.v.d., Li, Y., Vinyals, O.: Representation learning with contrastive
  predictive coding. arXiv preprint arXiv:1807.03748  (2018)

\bibitem{pareja2020evolvegcn}
Pareja, A., Domeniconi, G., Chen, J., Ma, T., Suzumura, T., Kanezashi, H.,
  Kaler, T., Schardl, T.B., Leiserson, C.E.: Evolvegcn: Evolving graph
  convolutional networks for dynamic graphs. In: Proceedings of the AAAI
  Conference on Artificial Intelligence. vol.~34, pp. 5363--5370. AAAI Press
  (2020)

\bibitem{poursafaei2022edgebank}
Poursafaei, F., Huang, S., Pelrine, K., Rabbany, R.: Towards better evaluation
  for dynamic link prediction. In: Advances in Neural Information Processing
  Systems. vol.~35, pp. 32928--32941 (2022)

\bibitem{rossi2020temporal}
Rossi, E., Chambers, B.P., Frasca, F., Eynard, D., Monti, F., Bronstein, M.:
  Temporal graph networks for deep learning on dynamic graphs. arXiv preprint
  arXiv:2006.10637  (2020)

\bibitem{sankar2020dysat}
Sankar, A., Wu, Y., Gou, L., Zhang, W., Yang, H.: Dysat: Deep neural
  representation learning on dynamic graphs via self-attention networks. In:
  Proceedings of the 13th International Conference on Web Search and Data
  Mining (WSDM). pp. 519--527. ACM (2020)

\bibitem{trivedi2019dyrep}
Trivedi, R., Farajtabar, M., Biswal, P., Zha, H.: Dyrep: Learning
  representations over dynamic graphs. In: International Conference on Learning
  Representations (ICLR) (2019)

\bibitem{vaswani2017attention}
Vaswani, A., Shazeer, N., Parmar, N., Uszkoreit, J., Jones, L., Gomez, A.N.,
  Kaiser, L., Polosukhin, I.: Attention is all you need. In: Advances in Neural
  Information Processing Systems. vol.~30 (2017)

\bibitem{wang2021tcl}
Wang, L., Chang, X., Li, S., Chu, Y., Li, H., Zhang, W., He, X., Song, L.,
  Zhou, J., Yang, H.: Tcl: Transformer-based dynamic graph modelling via
  contrastive learning. arXiv preprint arXiv:2105.07944  (2021)

\bibitem{wang2021cawn}
Wang, Y., Chang, Y.Y., Liu, Y., Leskovec, J., Li, P.: Inductive representation
  learning in temporal networks via causal anonymous walks. In: International
  Conference on Learning Representations (ICLR) (2021)

\bibitem{xu2020tgat}
Xu, D., Ruan, C., Korpeoglu, E., Kumar, S., Achan, K.: Inductive representation
  learning on temporal graphs. In: International Conference on Learning
  Representations (ICLR) (2020)

\bibitem{xu2023cldg}
Xu, Y., Shi, B., Ma, T., Dong, B., Zhou, H., Zheng, Q.: Cldg: Contrastive
  learning on dynamic graphs. In: 2023 IEEE 39th International Conference on
  Data Engineering (ICDE). pp. 696--707. IEEE (Apr 2023)

\bibitem{yu2023dygformer}
Yu, L., Sun, L., Du, B., Lv, W.: Towards better dynamic graph learning: New
  architecture and unified library. In: Advances in Neural Information
  Processing Systems (NeurIPS). vol.~36, pp. 67686--67700 (2023)

\bibitem{zhang2024dtgb}
Zhang, J., Chen, J., Yang, M., Feng, A., Liang, S., Shao, J., Ying, R.: Dtgb: A
  comprehensive benchmark for dynamic text-attributed graphs. In: Advances in
  Neural Information Processing Systems (NeurIPS). vol.~37, pp. 91405--91429.
  Curran Associates, Inc. (2024)

\bibitem{zhu2024idol}
Zhu, Z., Wang, K., Liu, H., Li, J., Luo, S.: Topology-monitorable contrastive
  learning on dynamic graphs. In: Proceedings of the 30th ACM SIGKDD Conference
  on Knowledge Discovery and Data Mining. pp. 4700--4711. ACM (Aug 2024)

\end{thebibliography}

\end{document}